\documentclass[11pt,a4paper]{article}
\usepackage{acl2000,amsmath,psfig}

\newcommand{\Reals}{\mathrm{I} \! \mathrm{R}}
\newcommand{\RpowN}[1]{{\mathrm{I} \! \mathrm{R}}^{#1}}

\title{Lexicalized Stochastic Modeling of Constraint-Based Grammars using Log-Linear Measures and EM Training}

\author{Stefan Riezler \\
IMS, Universit\"at Stuttgart \\
\texttt{\normalsize riezler@ims.uni-stuttgart.de}
\And
Detlef Prescher\\
IMS, Universit\"at Stuttgart \\
\texttt{\normalsize prescher@ims.uni-stuttgart.de}
\AND
Jonas Kuhn\\
IMS, Universit\"at Stuttgart \\
\texttt{\normalsize jonas@ims.uni-stuttgart.de}
\And
Mark Johnson\\
Cog. \& Ling. Sciences, Brown University \\
\texttt{\normalsize Mark\_Johnson@brown.edu}
}

\begin{document}

\maketitle

\begin{abstract}
We present a new approach to stochastic modeling of constraint-based grammars that
is based on log-linear models and uses EM for estimation
from unannotated data. The techniques are applied to an LFG
grammar for German. Evaluation on an exact match task yields 86\% precision
for an ambiguity rate of 5.4, and 90\% precision on a subcat frame match for 
an ambiguity rate of 25. Experimental comparison to training from a parsebank
shows a 10\% gain from EM training. Also, a new class-based grammar
lexicalization is presented, showing a 10\% gain over unlexicalized models.
\end{abstract}

\section{Introduction} 
\label{secintro}

Stochastic parsing models capturing contextual constraints beyond
the dependencies of probabilistic context-free grammars (PCFGs) are
currently the subject of intensive research. An interesting feature
common to most such models is the incorporation of contextual dependencies on
individual head words into rule-based probability models. Such
word-based lexicalizations of probability models are used
successfully in the statistical parsing models of, e.g.,
\newcite{Collins:97}, \newcite{Charniak:97}, or \newcite{Rat:97}.
However, it is still an open question which kind of lexicalization,
e.g., statistics on individual words or statistics based upon word
classes, is the best choice.
Secondly, these approaches have in common the fact that the probability models are trained on
treebanks, i.e., corpora of manually disambiguated sentences, and not
from corpora of unannotated sentences. In all of the cited approaches,
the Penn Wall Street Journal Treebank \cite{Marcus:93} is used, the
availability of which obviates the standard effort required for treebank
training---hand-annotating large corpora of specific domains of
specific languages with specific parse types. Moreover, common wisdom
is that training from unannotated  data via the expectation-maximization (EM) algorithm
\cite{Dempster:77} yields poor results unless at least partial
annotation is applied. Experimental results confirming this wisdom
have been presented, e.g., by \newcite{Elworthy:94} and
\newcite{Pereira:92} for EM training of Hidden Markov Models and
PCFGs.

In this paper, we present a new lexicalized stochastic model for 
constraint-based grammars that employs a combination of head-word
frequencies and EM-based clustering for grammar lexicalization.
Furthermore, we make crucial use of EM for estimating the 
parameters of the stochastic grammar from unannotated data. 
Our usage of EM was initiated by the current lack of large
unification-based treebanks for German. However, our
experimental results also show an exception to the common wisdom of the
insufficiency of EM for highly accurate statistical modeling.

Our approach to lexicalized stochastic modeling is based on the
parametric family of log-linear probability models, which is used to
define a probability distribution on the parses of a
Lexical-Functional Grammar (LFG) for German. In previous work on log-linear models for LFG by \newcite{Johnson:99}, pseudo-likelihood
estimation from annotated corpora has been introduced and experimented
with on a small scale. However, to our knowledge, to date no large LFG
annotated corpora of unrestricted German text are available. Fortunately, algorithms exist for statistical inference of log-linear
models from unannotated data \cite{RiezlerAIMS:99}. We apply 
this algorithm to estimate log-linear LFG models from large
corpora of newspaper text. In our largest experiment, we used 250,000
parses which were produced by parsing 36,000 newspaper sentences with
the German LFG. Experimental evaluation of our models on an
exact-match task (i.e. percentage of exact match of most probable parse
with correct parse) on 550 manually examined examples with on average 5.4 analyses gave
86\% precision. Another evaluation on a verb
frame recognition task (i.e. percentage of agreement between subcategorization
frames of main verb of most probable parse and correct parse) gave 90\%
precision on 375 manually disambiguated examples with an average
ambiguity of  25.
Clearly, a direct comparison of these results to state-of-the-art
statistical parsers cannot be made because of different training and
test data and other evaluation measures. However, we would like to
draw the following conclusions from our experiments:

\begin{itemize}
\item The problem of chaotic convergence behaviour of EM estimation
  can be solved for log-linear models. 
\item EM does help constraint-based grammars, e.g. using
  about 10 times more sentences and about 100 times more parses for EM
  training than for training from an automatically constructed
  parsebank can improve precision by about 10\%.
\item Class-based lexicalization can yield a gain in precision of about 10\%.
\end{itemize}

In the rest of this paper we introduce incomplete-data estimation for
log-linear models (Sec. \ref{secIM}), and present the actual design of our models (Sec. \ref{secprops}) and report our experimental results (Sec. \ref{secexps}).

\section{Incomplete-Data Estimation for Log-Linear Models}
\label{secIM}

\subsection{Log-Linear Models}

A log-linear distribution $p_\lambda(x)$ on the set of analyses $\mathcal{X}$ of a
constraint-based grammar can be defined as follows:
\[
p_\lambda(x) =
{Z_\lambda}^{-1}
e^{\lambda \cdot \nu(x)} p_0(x) 
\]
where $Z_\lambda = \sum_{x \in \mathcal{X}} 
e^{\lambda \cdot \nu(x)} p_0(x)$
is a normalizing constant, 
 $\lambda = (\lambda_1, \ldots, \lambda_n) \in \RpowN{n}$
is a vector of log-parameters, 
 $\mathbf{\nu} = (\nu_1, \ldots, \nu_n)$
is a vector of property-functions $\nu_i:\mathcal{X}
\rightarrow \Reals$ for $i= 1, \ldots, n$, 
 $\lambda \cdot \nu (x)$
is the vector dot product $\sum^n_{i=1}\lambda_i \nu_i(x)$,
and $p_0$ is a fixed reference distribution.

The task of probabilistic modeling with log-linear distributions
is to build salient properties of the data as property-functions
$\nu_i$ into the probability model. For a given vector $\nu$ of
property-functions, the task of statistical inference is to tune
the parameters $\lambda$ to best reflect the empirical distribution of
the training data.

\subsection{Incomplete-Data Estimation}

\begin{figure*}[htbp]
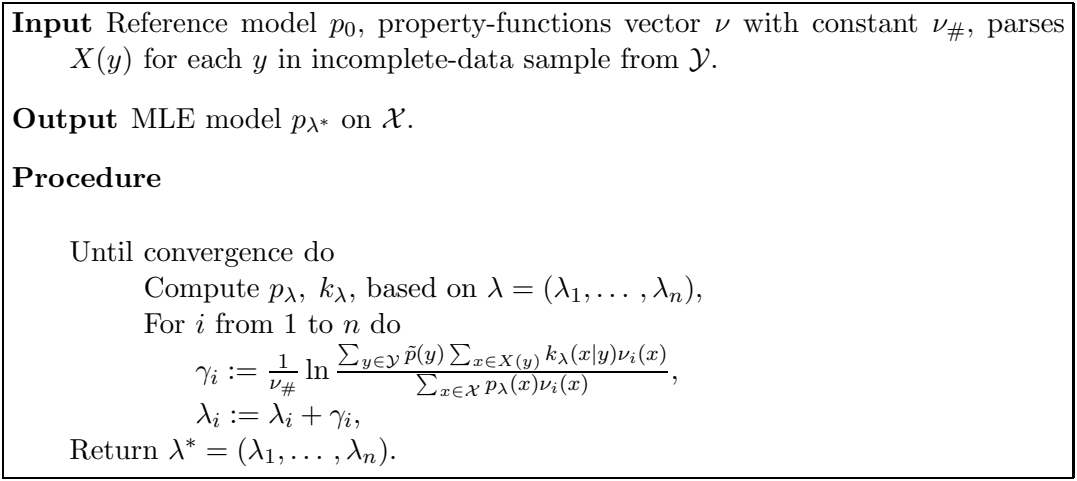
 
\begin{center}
\fbox{\begin{minipage}{14cm}

\begin{description}
\item[Input] Reference model $p_0$, 
  property-functions vector $\nu$ with constant $\nu_\#$,
 parses $X(y)$ for each $y$ in incomplete-data sample from $\mathcal{Y}$.

\item[Output] MLE model $p_{\lambda^\ast}$ on $\mathcal{X}$.

\item[Procedure] \mbox{}

\begin{tabbing}
Until \= convergence do \\
\> Compute $p_\lambda, \; k_\lambda$, based on $\lambda = (\lambda_1,
\ldots, \lambda_n)$, \\
\> For \= $i$ from $1$ to $n$ do \\
\>\> $\gamma_i := 
\frac{1}{\nu_\#} \ln
\frac{\sum_{y \in \mathcal{Y}} \tilde p(y) 
\sum_{x \in X(y)} 
k_{\lambda} (x|y) \nu_i(x)}
{\sum_{x \in \mathcal{X}} p_{\lambda} (x) \nu_i (x)} $, \\
\>\> $\lambda_i := \lambda_i + \gamma_i$, \\
Return $\lambda^\ast = (\lambda_1, \ldots, \lambda_n)$.
\end{tabbing}
\end{description}

\end{minipage}}
\end{center}
\caption{Closed-form version of IM algorithm}
\label{algo}
\end{figure*}

Standard numerical methods for statistical inference
of log-linear models from fully annotated data---so-called complete
data---are the iterative scaling methods of \newcite{Darroch:72} and
\newcite{DDL:97}. For data consisting of unannotated
sentences---so-called incomplete data---the iterative method of the EM
algorithm \cite{Dempster:77} has to be employed. However, since 
even complete-data estimation for log-linear models requires iterative methods, an application of EM to log-linear models results in an
algorithm which is expensive since it is doubly-iterative. A
singly-iterative algorithm interleaving EM and iterative scaling into a mathematically
well-defined estimation method for log-linear models from incomplete
data is the IM algorithm of \newcite{RiezlerAIMS:99}. Applying this algorithm to
stochastic constraint-based grammars, we assume the
following to be given: A training sample of
unannotated sentences $y$ from a set $\mathcal{Y}$, observed with
empirical probability $\tilde p(y)$, a constraint-based grammar yielding a set $X(y)$ of
parses for each sentence $y$, and a log-linear model
$p_\lambda(\cdot)$ on the parses $\mathcal{X}=\sum_{y \in
  \mathcal{Y}| \tilde p(y) > 0} X(y)$ for the sentences in the
training corpus, with known values of property-functions $\nu$ and unknown values
of $\lambda$. The aim of incomplete-data maximum likelihood
estimation (MLE) is to find a value $\lambda^\ast$ that maximizes the
incomplete-data log-likelihood $L = \sum_{y \in
  \mathcal{Y}} \tilde p(y) \ln \sum_{x \in X(y)} p_\lambda(x)$, i.e., 
\[
\lambda^\ast  =  \underset{\lambda \in \RpowN{n}}{\arg\max\;}
L(\lambda).
\]
Closed-form parameter-updates for this problem can be computed by the
algorithm of Fig. \ref{algo}, where $\nu_\#(x) = \sum_{i=1}^n
\nu_i(x)$, and $k_\lambda(x|y) = p_\lambda(x) /
\sum_{x \in X(y)} p_\lambda(x)$ is the conditional probability of a
parse $x$ given the sentence $y$ and the current
parameter value $\lambda$. 

The constancy requirement on $\nu_\#$ can be enforced by adding a ``correction'' property-function $\nu_l$: 
\begin{quote}
Choose $K = \max_{x \in \mathcal{X}} \: \nu_\#(x)$ and
$\nu_l(x) = K - \nu_\#(x)$ for all $x \in \mathcal{X}$. \\
Then $\sum^l_{i=1} \nu_i(x) = K$ for all $x \in \mathcal{X}$.
\end{quote}
Note that because of the restriction of $\mathcal{X}$ to the
parses obtainable by a grammar from the training corpus, we have a
log-linear probability measure only on those parses and not on all
possible parses of the grammar. We shall therefore speak of mere
log-linear measures in our application of disambiguation.

\subsection{Searching for Order in Chaos}

For incomplete-data estimation, a sequence of likelihood values is
guaranteed to converge to a critical point of the likelihood function
$L$. This is shown for the IM algorithm in \newcite{RiezlerAIMS:99}.
 The process of finding likelihood maxima is chaotic in that the final
 likelihood value is extremely sensitive to the starting values of
 $\lambda$, i.e. limit points can be local maxima (or saddlepoints),
 which are not necessarily also global maxima. A way to search for
order in this chaos is to search for starting values which are
hopefully attracted by the global maximum of $L$. This problem can best be
explained in terms of the minimum divergence paradigm
\cite{Kullback:59}, which is equivalent to the maximum likelihood
paradigm by the following theorem. Let $p[f] = \sum_{x \in
  \mathcal{X}} p(x) f(x)$ be the expectation of a function $f$
with respect to a distribution $p$:
\begin{quote}
The probability distribution $p^\ast$ that minimizes the divergence
$D(p||p_0)$ to a reference model $p_0$
subject to the constraints $p[\nu_i] = q[\nu_i], i = 1, \ldots,
n$ is the model in the parametric family of log-linear distributions
$p_\lambda$ that maximizes the likelihood $L(\lambda) = q[ \ln
p_\lambda ]$ of the training data\footnote{If the training sample
  consists of complete data $x \in \mathcal{X}$, the expectation
  $q[\cdot]$ corresponds to the empirical expectation $\tilde
  p[\cdot]$. If we observe incomplete data $y \in \mathcal{Y}$, the expectation $q[\cdot]$
  is replaced by the conditional expectation $\tilde p[
  k_{\lambda'}[\cdot]]$ given the observed data $y$ and the current
  parameter value $\lambda'$.}.
\end{quote}
Reasonable starting values for minimum divergence estimation
is to set $\lambda_i = 0$ for $i = 1, \ldots, n$. This yields a distribution which minimizes
the divergence to $p_0$, over the set of models $p$ to which the constraints $p[\nu_i] =
q[\nu_i], i = 1, \ldots, n$ have yet to be applied.
Clearly, this argument applies to both complete-data and
incomplete-data estimation. Note that for a uniformly distributed
reference model $p_0$, the minimum divergence model is a maximum
entropy model \cite{Jaynes:57}. In Sec. \ref{secexps}, we will demonstrate that a
uniform initialization of the IM algorithm shows a significant
improvement in likelihood maximization as well as in linguistic
performance when compared to standard random initialization.

\section{Property Design and Lexicalization}
\label{secprops}

\subsection{Basic Configurational Properties} \label{basicprops}

The basic 190 properties employed in our models are similar to the
properties of \newcite{Johnson:99} which incorporate general linguistic
principles into a log-linear model. They refer to both the
c(onstituent)-structure and the f(eature)-structure of the LFG parses.
Examples are properties for
\begin{itemize}
\setlength{\parsep}{0ex}
\setlength{\itemsep}{0ex}
\item c-structure nodes, corresponding to standard
  production properties,
\item c-structure subtrees, indicating argument versus
  adjunct attachment,
\item f-structure attributes, corresponding to
  grammatical functions used in LFG,
\item atomic attribute-value pairs in f-structures,
\item complexity of the phrase being
  attached to, thus indicating both high and low attachment, 
\item non-right-branching behavior of nonterminal nodes,
\item non-parallelism of coordinations.
\end{itemize}

\subsection{Class-Based Lexicalization}
\label{lexprops}

Our approach to grammar lexicalization is class-based in the
sense that we use class-based estimated frequencies $f_c(v,n)$ of
head-verbs $v$ and argument head-nouns $n$ instead of pure frequency
statistics or class-based probabilities of head word dependencies. Class-based estimated frequencies are
introduced in \newcite{Prescher:00} as the frequency
$f(v,n)$ of a $(v,n)$-pair in the training corpus, weighted by the
best estimate of the class-membership probability $p(c|v,n)$ of an EM-based clustering
model on $(v,n)$-pairs, i.e., $f_c(v,n) = \underset{c
  \in C}{\max\;}p(c|v,n) (f(v,n) +1)$. As is shown in
\newcite{Prescher:00} in an evaluation on lexical ambiguity
resolution, a gain of about 7\% can be obtained by using the
class-based estimated frequency 
$f_c(v,n)$ as disambiguation criterion instead of class-based
probabilities $p(n|v)$. In order to make the most direct use possible
of this fact, we incorporated
the decisions of the disambiguator directly into 45 additional
properties for the grammatical relations of the subject, direct object, indirect
object, infinitival object, oblique and adjunctival dative and
accusative preposition, for active and passive forms of the first
three verbs in each parse. Let $ v_r(x)$ be the verbal head of
grammatical relation $r$ in parse $x$, and $n_r(x)$ the nominal head
of grammatical relation $r$ in $x$. Then a lexicalized property
$\nu_r$ for grammatical relation $r$ is defined as
\[
\nu_r(x) = 
\left\{ 
\begin{array}{ll} 
1 &
\begin{array}{ll}
\textrm{if } f_c(v_r(x),n_r(x)) \geq \\
f_c(v_r(x'),n_r(x')) \; \forall x' \in X(y),
\end{array}\\
0 & \textrm{otherwise}.
\end{array} 
\right. 
\]
The property-function $\nu_r$ thus pre-disambiguates
the parses $x \in X(y)$ of a sentence $y$ according to $f_c(v,n)$, and stores
the best parse directly instead of taking the actual estimated
frequencies as its value. In Sec. \ref{secexps}, we will see that an
incorporation of this pre-disambiguation routine into the models improves performance
in disambiguation by about 10\%.

\section{Experiments}
\label{secexps}

\begin{figure*}[htbp]
\begin{center}
\begin{tabular}{|l|c|c|c|}
\hline
\begin{tabular}{c} \textbf{exact match} \\ \textbf{evaluation} \end{tabular}
& \begin{tabular}{c} basic \\ model  \end{tabular} 
& \begin{tabular}{c} lexicalized \\ model  \end{tabular} 
& \begin{tabular}{c} selected  \\ + lexicalized \\ model \end{tabular} \\ 
\hline
\begin{tabular}{c} complete-data \\ estimation  \end{tabular}
& \begin{tabular}{l} P: 68 \\ E: 59.6 \end{tabular} 
& \begin{tabular}{l} P: 73.9 \\ E: 71.6 \end{tabular} 
& \begin{tabular}{l} P: 74.3 \\ E: 71.8 \end{tabular} \\ \hline
\begin{tabular}{c} incomplete-data \\ estimation  \end{tabular}
& \begin{tabular}{l} P: 73 \\ E: 65.4 \end{tabular} 
& \begin{tabular}{l} P: 86 \\ E: 85.2 \end{tabular} 
& \begin{tabular}{l} P: 86.1 \\ E: 85.4 \end{tabular} \\ \hline
\end{tabular}
\end{center}
\caption{Evaluation on exact match task for 550 examples with average
  ambiguity 5.4}
\label{figexacteval}
\end{figure*}

\begin{figure*}[htbp]
\begin{center}
\begin{tabular}{|l|c|c|c|}
\hline
\begin{tabular}{c} \textbf{frame match}  \\ \textbf{evaluation} \end{tabular}
& \begin{tabular}{c} basic  \\ model \end{tabular}
& \begin{tabular}{c} lexicalized  \\ model \end{tabular}
& \begin{tabular}{c} selected  \\ + lexicalized \\ model \end{tabular} \\ 
\hline
\begin{tabular}{c} complete-data \\ estimation \end{tabular}
& \begin{tabular}{l} P: 80.6 \\ E: 70.4 \end{tabular} 
& \begin{tabular}{l} P: 82.7 \\ E: 76.4 \end{tabular} 
& \begin{tabular}{l} P: 83.4 \\ E: 76 \end{tabular} \\ \hline
\begin{tabular}{c} incomplete-data \\ estimation \end{tabular}
& \begin{tabular}{l} P: 84.5 \\ E: 73.1 \end{tabular} 
& \begin{tabular}{l} P: 88.5 \\ E: 84.9 \end{tabular} 
& \begin{tabular}{l} P: 90 \\ E: 86.3 \end{tabular} \\ \hline
\end{tabular}
\end{center}
\caption{Evaluation on frame match task for
  375 examples with average ambiguity 25}
\label{figparteval}
\end{figure*}

\subsection{Incomplete Data and Parsebanks}

In our experiments, we used an LFG grammar for German\footnote{The
  German LFG grammar is being implemented in the Xerox Linguistic Environment (XLE, see \newcite{MaxKap:96}) as part of the Parallel Grammar (ParGram) project at the IMS
  Stuttgart. The coverage of the grammar is about 50\% for
   unrestricted newspaper text. For the experiments reported here, the
   effective coverage was lower, since the corpus preprocessing we
   applied was minimal. Note that for the disambiguation task we were
   interested in, the overall grammar coverage was of subordinate
   relevance.} for parsing unrestricted text. Since training was faster than
parsing, we parsed in advance and stored the resulting packed
c/f-structures. The low ambiguity rate of the German LFG 
grammar allowed us to restrict the training data to
  sentences with at most 20 parses. The resulting training corpus of
  unannotated, incomplete data consists of approximately 36,000
  sentences of online available German newspaper text, comprising
  approximately 250,000 parses.  

In order to compare the contribution of unambiguous and
ambiguous sentences to the estimation results, we extracted a subcorpus
of 4,000 sentences, for which the LFG grammar produced a unique parse,
from the full training corpus. The average sentence length of 7.5 for this
automatically constructed parsebank is only slightly smaller than that of
10.5 for the full set of 36,000 training sentences and 250,000 parses.
Thus, we conjecture that the parsebank includes a representative
variety of linguistic phenomena. Estimation from this automatically
disambiguated parsebank enjoys the same complete-data estimation
properties\footnote{For example, convergence to the global maximum of
  the complete-data log-likelihood function is guaranteed, which is a
  good condition for highly precise statistical disambiguation.} as
training from manually disambiguated treebanks. This makes a
comparison of complete-data estimation from this parsebank to
incomplete-data estimation from the full set of training data interesting. 

\subsection{Test Data and Evaluation Tasks}

To evaluate our models, we constructed two different test corpora. We
first parsed with the LFG grammar 550 sentences which are used for illustrative
purposes in the foreign language learner's grammar of
\newcite{HelbigBuscha:96}. In a next step, the correct parse was
indicated by a human disambiguator, according to the reading intended
in \newcite{HelbigBuscha:96}. Thus a precise indication of correct
c/f-structure pairs was possible. However, the average ambiguity of
this corpus is only 5.4 parses per sentence, for sentences with on
average 7.5 words. In order to evaluate on sentences with higher
ambiguity rate, we manually disambiguated further 375 sentences of
LFG-parsed newspaper text. The sentences of this corpus have on average
25 parses and 11.2 words.

We tested our models on two evaluation tasks. The statistical
disambiguator was tested on an ``exact match'' task, where exact
correspondence of the full c/f-structure pair of the hand-annotated
correct parse and the most probable parse is checked. Another
evaluation was done on a ``frame match'' task, where exact correspondence only
of the subcategorization frame of the main verb of the most probable parse
and the correct parse is checked. Clearly, the latter task involves a
smaller effective ambiguity rate, and is thus to be interpreted as an
evaluation of the combined system of highly-constrained symbolic
parsing and statistical disambiguation.

Performance on these two evaluation tasks was assessed according to
the following evaluation measures:
\begin{center}
$\textrm{Precision} = \frac{\# \textrm{correct}} {\#
  \textrm{correct} + \# \textrm{incorrect}}$, \\ 
$\textrm{Effectiveness} = \frac{\# \textrm{correct}} {\#
  \textrm{correct} + \# \textrm{incorrect} + \# \textrm{don't know}}$.
\end{center}
``Correct'' and ``incorrect'' specifies a success/failure on the respective
evaluation tasks; ``don't know'' cases are cases where the system is
unable to make a decision, i.e. cases with more than one most probable
parse. 

\subsection{Experimental Results}

For each task and each test corpus, we calculated a random baseline by
averaging over several models with randomly chosen parameter values.
This baseline measures the disambiguation power of the pure symbolic parser.
The results of an exact-match evaluation on the Helbig-Buscha corpus is shown in Fig.
\ref{figexacteval}. The random baseline was around 33\% for this case. The columns
list different models according to their property-vectors. ``Basic''
models consist of 190 configurational properties as described in Sec.
\ref{basicprops}. ``Lexicalized'' models are extended by 45 lexical
pre-disambiguation properties as described in Sec. \ref{lexprops}.
``Selected + lexicalized'' models result from a simple property
selection procedure where a cutoff on the number of parses with
non-negative value of the property-functions was set. Estimation of
basic models from complete data gave 68\% precision (P), whereas training
lexicalized and selected models from incomplete data gave 86.1\% precision, which is
an improvement of 18\%. Comparing lexicalized models in the
estimation method shows that incomplete-data estimation gives an
improvement of 12\% precision over training from the parsebank. A comparison of
models trained from incomplete data shows that lexicalization yields a
gain of 13\% in precision. Note also the gain in effectiveness (E) due to the
pre-disambigution routine included in the lexicalized properties. The
gain due to property selection both in precision and effectiveness is
minimal. A similar pattern of performance arises in an exact match
evaluation on the newspaper corpus with an ambiguity rate of 25. The
lexicalized and selected model trained from incomplete data achieved
here 60.1\% precision and 57.9\% effectiveness, for a random baseline
of around 17\%.

As shown in Fig. \ref{figparteval}, the improvement in performance due
to both lexicalization and EM training is smaller for the easier task
of frame evaluation. Here the random baseline is 70\% for frame
evaluation on the newspaper corpus with an ambiguity rate of 25. An overall
gain of roughly 10\% can be achieved by going from unlexicalized
parsebank models (80.6\% precision) to lexicalized EM-trained models
(90\% precision). Again, the contribution to this improvement is about
the same for lexicalization and incomplete-data training. Applying the same
evaluation to the Helbig-Buscha corpus shows 97.6\% precision and
96.7\% effectiveness for the lexicalized and selected incomplete-data
model, compared to around 80\% for the random baseline.

\begin{figure*}
\begin{center}
\mbox{\psfig{file=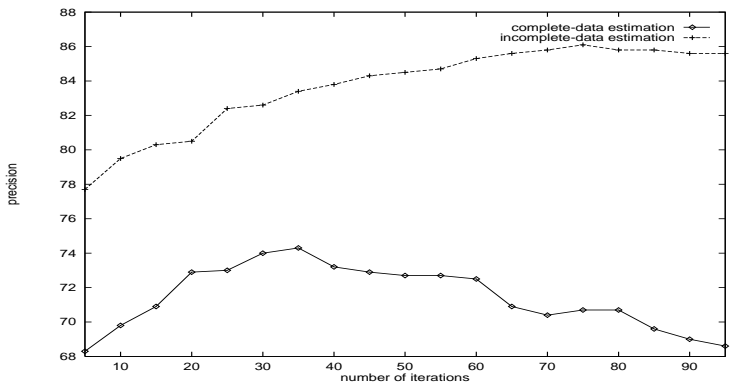,angle=270,width=10cm,height=5cm}}
\end{center}
\caption{Precision on exact match task in number of training iterations}
\label{figiters}
\end{figure*}

Optimal iteration numbers were decided by repeated evaluation of the models at
every fifth iteration. Fig. \ref{figiters} shows the precision of
lexicalized and selected models on the exact match task plotted
against the number of iterations of the training algorithm. For
parsebank training, the maximal precision value is obtained at 35
iterations. Iterating further shows a clear overtraining effect. For
incomplete-data estimation more iterations are necessary to reach a
maximal precision value.
A comparison of models with random or uniform starting values shows an
increase in precision of 10\% to 40\% for the latter. In terms of
maximization of likelihood, this corresponds to the fact that uniform
starting values immediately push the likelihood up to nearly its final value, whereas random
starting values yield an initial likelihood which has to be increased
by factors of 2 to 20 to an often lower final value.

\section{Discussion}
\label{secdisc}

The most direct points of comparison of our method are the approaches of
\newcite{Johnson:99} and \newcite{Johnson:00}. In the first approach,
log-linear models on LFG grammars using about 200 configurational
properties were trained on treebanks of about 400 sentences by maximum
pseudo-likelihood estimation. Precision was evaluated on an exact match task in a 10-way cross
validation paradigm for an ambiguity rate of 10, and achieved 59\% for the
first approach. \newcite{Johnson:00} achieved a gain of 1\% over
this result by including a class-based lexicalization. Our best models
clearly outperform these results, both in terms of precision relative
to ambiguity and in terms of relative gain due to lexicalization.
A comparison of performance is more difficult for the lexicalized PCFG
of \newcite{Beil:99} which was trained by EM on 450,000 sentences of
German newspaper text. There, a 70.4\% precision is reported on a verb frame recognition task
on 584 examples. However, the gain
achieved by \newcite{Beil:99} due to grammar lexicalizaton 
is only 2\%, compared to about 10\% in our case. 
A comparison is difficult also for most other
state-of-the-art PCFG-based statistical parsers, since different
training and test data, and most importantly, different evaluation
criteria were used. A comparison of the performance gain due to
grammar lexicalization shows that our results are on a par with that
reported in \newcite{Charniak:97}.

\section{Conclusion}
\label{secconc}

We have presented a new approach to stochastic modeling of
constraint-based grammars. Our experimental results show that EM
training can in fact be very helpful for accurate stochastic modeling
in natural language processing. We conjecture that this result is due
partly to the fact that the space of parses produced by a
constraint-based grammar is only ``mildly incomplete'', i.e. the
ambiguity rate can be kept relatively low. Another reason may be that EM
is especially useful for log-linear models, where the search space in
maximization can be kept under control. Furthermore, we have introduced a new
class-based grammar lexicalization, which again uses EM training and
incorporates a pre-disambiguation routine into log-linear models. An
impressive gain in performance could also be demonstrated for this
method.
Clearly, a central task of future work is a further exploration of
the relation between complete-data and incomplete-data estimation for
larger, manually disambiguated treebanks. An interesting question is
whether a systematic variation of training data size along the lines of the
EM-experiments of \newcite{Nigam:00} for text classification will
show similar results, namely a systematic dependence of the relative
gain due to EM training from the relative sizes of unannotated and
annotated data. Furthermore, it is important to show that EM-based methods
can be applied successfully also to other statistical parsing
frameworks.

\section*{Acknowledgements}

We thank Stefanie Dipper and Bettina Schrader for help
with disambiguation of the test suites, and the anonymous ACL
reviewers for helpful suggestions. This research was supported by the
ParGram project and the project B7 of the SFB 340 of the DFG.

\bibliographystyle{acl}

\end{document}